\documentclass[lettersize,journal]{IEEEtran}
\usepackage{amsmath,amsfonts,amssymb,mathrsfs}
\usepackage{algorithmic}
\usepackage{array}
\usepackage[caption=false,font=normalsize,labelfont=sf,textfont=sf]{subfig}
\usepackage{textcomp}
\usepackage{stfloats}
\usepackage{url}
\usepackage{verbatim}
\usepackage{graphicx}
\usepackage{booktabs}
\usepackage{tabularx}
\usepackage{xcolor}
\usepackage{hyperref}
\usepackage{nomencl}
\makenomenclature
\usepackage{multirow}
\usepackage{svg}

\usepackage{tikz}
\usetikzlibrary{shapes.geometric, positioning}
\usepackage{balance}

\usepackage{amsthm}  % 
\newtheorem{remark}{Remark}

\tikzstyle{startstop} = [
    rectangle, 
    rounded corners, 
    minimum width=1.4cm, 
    minimum height=0.8cm, 
    text centered, 
    draw=black, 
    fill=gray!5,
    align=center
]
\tikzstyle{decision} = [
    diamond, 
    aspect=2,
    text centered, 
    draw=black, 
    fill=gray!5,
    align=center
]
\tikzstyle{process} = [
    rectangle, 
    rounded corners, 
    minimum width=1.4cm, 
    minimum height=0.8cm, 
    text centered, 
    draw=black, 
    fill=gray!5,
    align=center
]
\tikzstyle{arrow} = [thick,->,>=stealth]

\begin{document}
\title{Performance and Experimental Analysis of Strain-based Models for Continuum Robots}
\author{Annika Delucchi, \and Vincenzo Di Paola, \and Andreas M\"{u}ller \and and Matteo Zoppi
 
\thanks{
Annika Delucchi is with the Department of Mechanical, Energy, Management and Transportation Engineering (DIME), University of Genova, Italy (e-mail: annika.delucchi@edu.unige.it)

Vincenzo Di Paola is with the Department of Mechanical, Energy, Management and Transportation Engineering (DIME), University of Genova, Italy (e-mail: vincenzo.dipaola@edu.unige.it).

Andreas M\"{u}ller is with the Institute of Robotics, Johannes Kepler University (JKU) Linz, Austria (e-mail: a.mueller@jku.at)

Matteo Zoppi is with the Department of Mechanical, Energy, Management and Transportation Engineering (DIME), University of Genova, Italy (e-mail: matteo.zoppi@unige.it).}}

\markboth{Preprint}%
{Shell \MakeLowercase{\textit{et al.}}: A Sample Article Using IEEEtran.cls for IEEE Journals}

\maketitle

\begin{abstract}

Although strain-based models have been widely adopted in robotics, no comparison beyond the uniform bending test is commonly recognized to assess their performance.
In addition, the increasing effort in prototyping continuum robots highlights the need to assess the applicability of these models and the necessity of comprehensive performance evaluation.
To address this gap, this work investigates the shape reconstruction abilities of a third-order strain interpolation method, examining its ability to capture both individual and combined deformation effects. These results are compared and discussed against the Geometric-Variable Strain approach. Subsequently, simulation results are experimentally verified by reshaping a slender rod while recording the resulting configurations using cameras. 
The rod configuration is imposed using a manipulator displacing one of its tips and extracted through reflective markers, without the aid of any other external sensor -- \textit{i.e.} strain gauges or wrench sensors placed along the rod.
The experiments demonstrate good agreement between the model predictions and observed shapes, with average error of $0.58\%$ of the rod length and average computational time of $0.32$\textnormal{s} per configuration, outperforming existing models.

\end{abstract}


\section{Introduction}

\IEEEPARstart{C}{ontinuum} robots are slender hyper elastic structures whose design and working mode are typically inspired to biological systems~\cite{KwiecinskiGoriely2017} and animals, as elephant trunks and octopus tentacles. Inheriting their morphological properties, these robots are well suited for grasping, manipulation and locomotion tasks while enabling delicate interactions with the surroundings. Among the many application areas one finds search-and-rescue operations~\cite{Wolf2003}, underwater and space orbital tasks~\cite{Ticozzi2025, Russo2023Space} and robotic-assisted surgery~\cite{BurgnerKahrs2015, Dupont2022}. Indeed, continuum robots can work as conventional medical tools, like flexible needles~\cite{Webster2006, Mahoney2016}, endoscopes~\cite{Ikuta1988} and catheters~\cite{Ohta2001}. Due to their intrinsic elasticity and control requirements~\cite{DellaSantina2023Review}, the modelling of these robots represents a unique challenge. Indeed, beyond achieving accuracy in shape reconstruction, computational efficiency is a must
%strict requirement 
for real-time applications, motivating research effort in developing improved models that strike this balance.
%. Researchers have focused on leveraging classical beam theories to develop sophisticated models that strike this balance. 

Continuum robots are often modelled as Cosserat rods, that can be thought as a stack of rigid cross-sections undergoing finite displacements.
The rod's configuration is then retrieved from the description of the cross-sectional poses, using either absolute or relative parametrisations.
The Finite Element Method (\textit{FEM}) is commonly used to discretise the rod reducing the system's order. The Geometrically Exact \textit{FEM} (\textit{GE-FEM}) introduced by Simo~\cite{Simo1985_Pt2} extends the \textit{FEM} concepts by considering the orientation to get configuration exactness.
However, the distinct interpolation between displacements and rotations, that are coupled quantities, limits the accuracy and robustness of the method~\cite{Cesarek2013}. To address this issue, different strain-based formulation were proposed so far. The Piece-wise Constant Strains (\textit{PCS}) method~\cite{Renda2018PCS} extends the geometric Piece-wise Constant Curvature (\textit{PCC}) approximation to mechanical models. Subsequently, integrating this concept with Cosserat produces a strain-based \textit{GE-FEM} model where cross-sectional frames are related through constant deformation measures.
General variable-strain formulations were thus proposed in~\cite{Zupan2003} for statics and in~\cite{Cesarek2013} for dynamics. Working in the strain field allows avoiding the mentioned issue of rotation-displacement by exploiting the additivity of vector spaces to generate kinematically consistent transformations in $SE(3)$~\footnote{Special Euclidean group.}.
The Geometric Variable Strains (\textit{GVS}) method~\cite{Boyer2019IEEE} models strain variations using arbitrary shape functions leveraging the Ritz method. The result is a reduced Lagrangian formulation in which the continuous spatial problem is approximated by a chosen functional basis for the strain field.

Although the promising performances in reduced-order models are interesting~\cite{Armanini2023, Russo2023}, there is a lack in experimental investigation. A measure of the performance is given through shape reconstruction experiments, that is by comparing predicted and experimental rod shapes for prescribed boundary conditions.  
In this context, stochastic shape estimation~\cite{BarfootBook_StateEstimation} inspired~\cite{Lilge2022} in using Gaussian regression to estimate the robot's shape. Subsequently,~\cite{FergusonRuckerWebster2024} incorporated actuation variables and external forces.
Further contributions presented estimators based on Cosserat theory to recover both material and space variables without the aid of external motion capture systems. The work of~\cite{ZhengBurgner2025} exploited the duality between tip velocities and internal wrenches to develop a dynamic state estimator, while~\cite{FeliuTalegonRenda2025} used only actuation measurements, that is cables length and tension, to retrieve both shape and external forces on tendon-driven continuum robots.

The shape reconstruction problem for the dual manipulation of continuum slender elements, commonly referred to as Deformable Linear Objects (\textit{DLOs}), is formulated in direct analogy.
The main difference is that the rod is constrained by prescribed end-tip poses, along with the absence of actuation units.
Several works have explored hybrid model-based and learned-based approaches to accomplish manipulation. 
The work of~\cite{Yu2023} proposes a method for learning a global deformation model
by, first, offline training a neural network on random data, and, then updating it online during manipulation tasks.
In~\cite{CaporaliPalli2025} a neural network predictive model is combined with a discrete Cosserat model to enable reliable \textit{DLO} tracking during manipulation. 
Nonetheless, a recurring limitation is that only a subset of the model’s state variables can be embedded into the neural network, losing accuracy and physical consistency, unless a fully physics-informed approach is adopted~\cite{Bensch2024}.
Moving to fully model-based approaches, the recent work~\cite{TiburzioDellaSantina2025} experimentally validates a strain-based polynomial curvature model for \textit{DLOs}, and demonstrates its feasibility for model-based control. In the latter, the experiments are conducted in planar settings and focused on shape regularization tasks. Despite the progress, there is still a lack of spatial testing of strain models and, in general of a standard procedure to assess the accuracy and computational time of them.

This work aims at testing the third-degree strain interpolated model presented in~\cite{Mueller2024}, both in simulation and through experiments. 
The simulations are set to show the main predictions of the presented model through comparison with the \textit{GVS}~\cite{Boyer2019IEEE}. Subsequently, spatial experiments are performed to confirm predicted results and model's features.
To date, these two approaches represent the most general reduced order strain-based models techniques in robotics.
Therefore, this work not only provides a one to one numerical comparison, but also a spatial experimental setup to test them through quasi-static dual-manipulation of a rod. 
The performance of the models is evaluated in terms of both solution accuracy and computational cost. In particular, the predicted rod shapes are compared against experimental measurements and numerical exact solutions. In addition, a metric is proposed to support the evaluation of strain-based model performance.

The paper is organized as follows. Section~\ref{sec:kin-of-rods} reviews the kinematics of Cosserat and Kirchhoff rod models. Section~\ref{sec:reduced-order-models} recalls the strain-parametrized models. Section~\ref{sec:evaluation-metrics} introduces the evaluation metric used to assess model performance, which is first applied in simulation in Section~\ref{sec:simulations} and subsequently evaluated experimentally in Section~\ref{sec:experiments}. Finally, Section~\ref{sec:conclusion} concludes the article.

\begin{table}[t]
    \centering
    \caption{Notation}
    \begin{tabularx}{\columnwidth}{
        >{\centering\arraybackslash}p{0.3\linewidth}
        >{\centering\arraybackslash}p{0.7\linewidth}
    }
        \toprule
        $L$ & Rod length \\
        $s$ & Curvilinear abscissa \\
        $\tau$ & Normalised curvilinear abscissa \\
        $\mathbf{R}(\tau)$ & Rotation matrix \\
        $\mathbf{r}(\tau)$ & Position vector \\
        $\mathbf{H}(\tau)$ & Transformation matrix \\
        $\mathbf{X}(\tau)$ & Canonical coordinates \\
        $\mathbf{x}(\tau)$ & Angular variation \\
        $\mathbf{y}(\tau)$ & Linear variation \\
        $\boldsymbol{\chi}(\tau)$ & Deformation \\
        $\bar{\boldsymbol{\chi}}(\tau)$ & Reference deformation \\
        $\boldsymbol{\chi}(\tau) - \bar{\boldsymbol{\chi}}(\tau)$ & Strain measure \\
        $\boldsymbol{\kappa}(\tau)$ & Bending and torsion strain \\
        $\boldsymbol{\rho}(\tau)$ & Shear and extension strain \\
        $\boldsymbol{\Lambda}(\tau)$ & Stress field \\
        % $\boldsymbol{\Lambda} (\tau)$ & External wrench \\
        $\mathbf{W}(\tau)$ & Distributed loads \\
        \bottomrule
    \end{tabularx}
    \label{table:list-of-symbols}
\end{table}

\section{Rod Model}
\label{sec:kin-of-rods}
This section introduces the nomenclature and the Kirchhoff equations starting from the Cosserat one. The notation used throughout the paper is summarised in Table~\ref{table:list-of-symbols}. The models are developed using Lie group theory, and the reader is referred to~\cite{Selig2004, MurrayLibro} and Appendix~\ref{app:Lie group} to review the main concepts and have explicit formulas of the functions employed in this work.

\subsection{Kinematics of Rods}
\begin{figure}[t]
    \centering
    \includegraphics[width=\columnwidth]{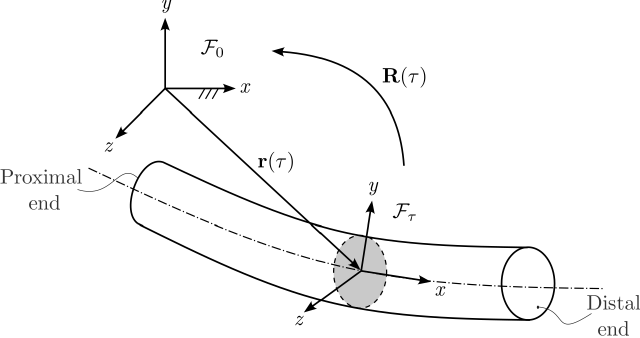}
    \caption{Rod description and parameters: a generic rod undergoing large deformations, where at each $\tau$ corresponds a reference frame $\mathcal{F}_{\tau}$ whose $x$-axis is aligned with the rod centreline.}
    \label{fig:rod}
\end{figure}

Consider a generic rod subject to large displacements and small deformations (Figure~\ref{fig:rod}). Call $L$ its length and $s$ the arc-length parameter.
Then, define the normalized arc-length parameter $\tau = s/L \in [0,1]$ 
parametrising the pose of the generic rigid cross-section as 
\begin{align*}
     \mathbf{H}(\tau) = \begin{pmatrix} \mathbf{R}(\tau) & \mathbf{r}(\tau) \\ \mathbf{0} & 1 \end{pmatrix} \in SE(3), \notag
\end{align*}
with $\mathbf{R}(\tau) \in SO(3)$ and $\mathbf{r}(\tau) \in \mathbb{R}^3$. 
Therefore, the rod configuration space $Q$ can be described in terms of its cross-sections as
\begin{align*}
Q := \big\{ & \mathbf{H}(\tau) \, \big| \, \tau \in [0,1], \, \mathbf{H}(\tau) \in SE(3) \big\}. \notag
\end{align*}
Notice that the cross-sectional pose can equivalently be retrieved through a vector of canonical coordinates $\mathbf{X}(\tau)$, containing an infinitesimal rotation $\mathbf{x}(\tau)$ and a translation $\mathbf{y}(\tau)$ and belonging to the Lie algebra $se(3)$ of $SE(3)$, 
\begin{align*}
    \mathbf{X}(\tau) = \begin{pmatrix}
        \mathbf{x}(\tau)^{\text{T}} & \mathbf{y}(\tau)^{\text{T}}
    \end{pmatrix}^{\text{T}}. 
\end{align*}
Indeed, the cross-sectional pose is obtained as
\begin{align}
    \mathbf{H}(\tau) = \mathbf{H}_0 \, \text{exp} \, \mathbf{X}(\tau),
    \label{eq:displacement-field}
\end{align}
with exp representing the exponential map defined as $\text{exp}~:~se(3)~\rightarrow~SE(3)$. 
The instantaneous motion of the cross-section yields the corresponding Lie algebra element encoding the local deformation of the cross-section
\begin{align}
\hat{\boldsymbol{\chi}}(\tau)
= \mathbf{H}^{-1}(\tau)\,\mathbf{H}'(\tau)
=
\begin{pmatrix}
\tilde{\boldsymbol{\kappa}}(\tau) & \boldsymbol{\rho}(\tau) \\
0 & 0
\end{pmatrix},
\label{eq: chi = H^-1 H'}
\end{align}
with $\hat{(\cdot)} \,: \mathbb{R}^6 \rightarrow \mathbb{R}^{4 \times 4}$, $\tilde{(\cdot)} \, : \mathbb{R}^3 \rightarrow \mathbb{R}^{3 \times 3}$ and $(\cdot)' = \frac{\partial}{\partial \tau}$. The deformation field $\boldsymbol{\chi}(\tau) \in se(3)$ of the cross-section contains, in general, torsion $\kappa_1$, bending $\kappa_2, \kappa_3$, elongation $\rho_1$ and shear $\rho_2, \rho_3$ modes.
A strain measure is defined as $\boldsymbol{\chi}(\tau) - \bar{\boldsymbol{\chi}}(\tau)$, where $\Bar{\boldsymbol{\chi}}(\tau)$ represents the reference deformation of the rod.

At this point, adopting the canonical coordinates representation of the displacement field in Eq.~\eqref{eq:displacement-field}, the kinematics of a Cosserat rod can be written as
% The \textcolor{red}{local} kinematic equations  
\begin{align}
    \boldsymbol{\chi}(\tau) = \mathbf{dexp}_{-\hat{\mathbf{X}}(\tau)} \mathbf{X}'(\tau),
    \label{eq:cosserat-kinematic-chi}
\end{align}
This is also known as local kinematic reconstruction equation~\cite{Mueller2021} and can be used to retrieve the cross-sectional pose in terms of the deformation field $\boldsymbol{\chi}(\tau)$ as
\begin{align}
    \mathbf{X}'(\tau) = \mathbf{dexp}_{\mathbf{X}(\tau)}^{-1} \boldsymbol{\chi}(\tau).
    \label{eq:cosserat-kinematic-X'}
\end{align}

\subsection{Constitutive equations}
Hooke's constitutive equation linearly relates the stress $\boldsymbol{\Lambda}(\tau)$ and strain $\boldsymbol{\chi}(\tau)$ 
\begin{align}
    \label{eq:cosserat-constitutive}
    \boldsymbol{\Lambda}(\tau) = \mathbf{K}(\tau) (\boldsymbol{\chi}(\tau) - \bar{\boldsymbol{\chi}}(\tau)),
\end{align}
where $\boldsymbol{\Lambda}$ belongs to $se(3)^*$, the dual of $se(3)$, while $\mathbf{K}$ represents the stiffness matrix,
\begin{align*}
    \mathbf{K}(\tau) &= \begin{pmatrix}
        \mathbf{K}_{BT}(\tau) & \mathbf{0} \\ \mathbf{0} & \mathbf{K}_{SE}(\tau)
    \end{pmatrix} \\ 
    &= \text{diag}(GI_{x}, \, EI_{y}, \, EI_{z}, \, EA, \, GA, \, GA),
\end{align*}
with $\mathbf{K}_{BT}(\tau), \,\mathbf{K}_{SE}(\tau) \in \mathbb{R}^{3\times3}$ representing the angular and the linear stiffness matrices respectively, $E$ Young modulus, $G$ shear modulus, $I$ moment of inertia and $A$ area of the cross-section.

\subsection{Equilibrium Equations}
Using the classical Euler-Poincaré equations one gets
\begin{align}
    \boldsymbol{\Lambda}'(\tau)- \mathbf{ad}_{\boldsymbol{\chi}(\tau)}^{\text{T}} {\boldsymbol{\Lambda}(\tau)} = \mathbf{W}(\tau),
    \label{eq:cosserat-euler-equation}
\end{align}
with $\mathbf{W}(\tau)$ being the distributed loads along the rod.
Combining Eq.~\eqref{eq:cosserat-constitutive} and Eq.~\eqref{eq:cosserat-euler-equation}, assuming constant properties of the material, and neglecting external loads yields the kinetostatic equilibrium
\begin{align}
    \begin{cases}
        \boldsymbol{\chi}'(\tau)= \bar{\boldsymbol{\chi}}'(\tau) + \mathbf{K}^{-1}\mathbf{ad}_{\boldsymbol{\chi}(\tau)}^{\text{T}} (\boldsymbol{\chi}(\tau) - \bar{\boldsymbol{\chi}}(\tau)) \\
        \mathbf{X}'(\tau) = \mathbf{dexp}_{-\hat{\mathbf{X}}(\tau)}^{-1} \boldsymbol{\chi}(\tau).
    \end{cases}
    \label{eq:cosserat-ODE}
\end{align}
This is the system of ODE to be solved to retrieve the rod's configuration in terms of its cross-section deformation.
% \begin{align*}
%     \boldsymbol{\chi}'(\tau)&= \bar{\boldsymbol{\chi}}'(\tau) + \mathbf{K}^{-1}(\tau)(\mathbf{ad}_{\boldsymbol{\chi}(\tau)}^{\text{T}}\mathbf{K}(\tau) - \mathbf{K}'(\tau))(\boldsymbol{\chi}(\tau) - \bar{\boldsymbol{\chi}}(\tau)) \notag \\
%     &+ \mathbf{K}^{-1}(\tau) \mathbf{W}(\tau).
% \end{align*}
% This equation can be simplified assuming constant properties of the material, that is $\mathbf{K}(\tau) =$ constant, and neglecting external loads, then
% \begin{align*}
%     \boldsymbol{\chi}'(\tau)= \bar{\boldsymbol{\chi}}(\tau) + \mathbf{K}^{-1}\mathbf{ad}_{\boldsymbol{\chi}(\tau)}^{\text{T}} (\boldsymbol{\chi}(\tau) - \bar{\boldsymbol{\chi}}(\tau)).
% \end{align*}
% The system of ODEs to be solved to retrieve the rod's configuration in terms of its cross-section deformation is

\begin{remark}
Neglecting certain components of the deformation vector $\boldsymbol{\chi}$ leads to models that allow less deformation modes. Some of the existing models are shown in Table~\ref{table:deformation-models}.
\end{remark}

Considering slender rods, whose cross-sectional diameter is much smaller than their length, made of inextensible materials, such as fiberglass or nitinol, allows neglecting shear and extensibility. Indeed, this work adopts the Kirchhoff inextensible model, where rod extension is not permitted, $\rho_1 = 1$, and the shear is neglected, $\rho_2 = \rho_3 = 0$. Practically, this is equivalent to enforcing the kinematic constraint $\boldsymbol{\rho}(\tau) = \begin{pmatrix} 1 & 0 & 0 \end{pmatrix}^{\text{T}} = \mathbf{e}_1$. Thus, the solutions of the equations \eqref{eq:cosserat-ODE} are constrained to a submanifold of $SE(3)$, where the kinematic constraint holds, and the updated equations for the inextensible Kirchhoff rod can be derived from Eq.~\eqref{eq:cosserat-kinematic-X'}. By imposing 
$\boldsymbol{\rho} = \mathbf{e}_1$ and substituting $\mathbf{y} = \bf{dexp}_{\hat{\mathbf{x}}}^{-1} \mathbf{r}$ coming from the definition of the exponential map, returns
\begin{align*}
    \mathbf{x}' = \bf{dexp}_{\mathbf{x}}^{-1} \boldsymbol{\kappa}, \hspace{1cm}
    \mathbf{r}' = \exp \, \mathbf{\tilde{x}} \, \mathbf{e}_1, 
\end{align*}
suggesting that position and orientation of the cross-section can be computed separately, as the latter depends on the curvature only. To be precise, the position can be integrated with quadrature to speed-up the computation. 
\begin{table}[]
    \centering
    \caption{Deformation Models}
    \begin{tabularx}{\columnwidth}{
        >{\centering\arraybackslash}p{0.3\linewidth}
        >{\centering\arraybackslash}p{0.7\linewidth}
    }
        \toprule
        MODEL & DEFORMATION \\
        \midrule
        Cosserat & $\boldsymbol{\chi} = \begin{pmatrix}
        \boldsymbol{\kappa}_1 & \boldsymbol{\kappa}_2 & \boldsymbol{\kappa}_3 & \boldsymbol{\rho}_1 & \boldsymbol{\rho}_2 & \boldsymbol{\rho}_3
        \end{pmatrix}^{\text{T}}$ \\
        Kirchhoff & $\boldsymbol{\chi} = \begin{pmatrix}
        \boldsymbol{\kappa}_1 & \boldsymbol{\kappa}_2 & \boldsymbol{\kappa}_3 &     \boldsymbol{\rho}_1 & 0 & 0
        \end{pmatrix}^{\text{T}}$ \\
        Inextensible Kirchhoff & $\boldsymbol{\chi} = \begin{pmatrix}
        \boldsymbol{\kappa}_1 & \boldsymbol{\kappa}_2 & \boldsymbol{\kappa}_3 & 1 & 0 & 0
        \end{pmatrix}^{\text{T}}$ \\
        Torsion Free Kirchhoff & $\boldsymbol{\chi} = \begin{pmatrix}
        0 & \boldsymbol{\kappa}_2 & \boldsymbol{\kappa}_3 &     \boldsymbol{\rho}_1 & 0 & 0
        \end{pmatrix}^{\text{T}}$ \\
        % Kirchhoff inextensible untorsionable & $\boldsymbol{\chi} = \begin{pmatrix}
        % 0 & \boldsymbol{\kappa}_2 & \boldsymbol{\kappa}_3 & 1 & 0 & 0
        % \end{pmatrix}^{\text{T}}$ \\
        \bottomrule
    \end{tabularx}
    \label{table:deformation-models}
\end{table}
Assuming the reference $\bar{\boldsymbol{\chi}} = \begin{pmatrix} 0 & 0 & 0 & 1 & 0 & 0 \end{pmatrix}$ and defining the constraint force $\boldsymbol{\mathscr{f}}=\mathbf{K}_{SE} \, (\boldsymbol{\rho} - \bar{\boldsymbol{\rho}})$ such that $\boldsymbol{\rho} = \mathbf{e}_1$ leads to 
\begin{align*}
    & \boldsymbol{\kappa}' = \mathbf{K}_{BT}^{-1} \, \tilde{\boldsymbol{\kappa}} \, \mathbf{K}_{BT}\boldsymbol{\kappa} +  \mathbf{K}_{BT}^{-1} \, \tilde{\mathbf{e}_1} \, \boldsymbol{\mathscr{f}} \\
    &  \boldsymbol{\mathscr{f}}' = - \tilde{\boldsymbol{\kappa}}  \boldsymbol{\mathscr{f}}.
\end{align*}
Observe that, when no external wrenches acts on the rod, the derivative of $\boldsymbol{\mathscr{f}}$ becomes constant, thus it can be  expressed in the reference frame $\mathcal{F}_0$ and propagated along the rod length.
Finally, the system of \textit{ODEs} to be solved becomes 
\begin{align}
    \begin{cases}
        \boldsymbol{\kappa}' = \mathbf{K}_{BT}^{-1} \, \tilde{\boldsymbol{\kappa}} \, \mathbf{K}_{BT}\boldsymbol{\kappa} +  \mathbf{K}_{BT}^{-1} \, \tilde{\mathbf{e}_1} \, \text{exp}(-\mathbf{x}) \, ^0\boldsymbol{\mathscr{f}} \\
        \mathbf{x}' = \bf{dexp}_{\mathbf{x}}^{-1} \boldsymbol{\kappa} \\
        \mathbf{r}' = \exp \, \mathbf{\tilde{x}} \, \mathbf{e}_1.
    \end{cases}
    \label{eq:kirchhoff-ODE}
\end{align}

\section{Approximated Strain-based Models}
\label{sec:reduced-order-models}
\subsection{Third-order Strain Interpolated Approach}
\label{sec:interp-model}
The Cosserat rod model can be analytically approximated to efficiently find solutions to the shape reconstruction problem. Indeed, for known initial and final poses and strains 
\begin{align*}
    &\mathbf{X}(0) = \mathbf{0} && \mathbf{X}(1) = \mathbf{X}_1  \notag \\
    &\mathbf{X}'(0) = \boldsymbol{\chi}(0) = \boldsymbol{\chi}_0, && \mathbf{X}(1)' = \mathbf{\bf{dexp}}_{-\mathbf{X}_1}^{-1} \boldsymbol{\chi}_1.
\end{align*}
one can obtain a third-order interpolation of the displacement fields as
\begin{align*}
    \mathbf{X}(\tau)^{[3]} &= (3\tau^2 - 2\tau^3)\mathbf{X}_1 + \tau(\tau - 1)^2 \boldsymbol{\chi}_0 \notag \\& + (\tau^3 - \tau^2) \mathbf{dexp}^{-1}_{-\mathbf{X}_1} \boldsymbol{\chi}_1.
\end{align*}
Substituting the derivative of $ \mathbf{X}(\tau)^{[3]}$ into Eq.~\eqref{eq:cosserat-kinematic-chi}, the deformation field $\boldsymbol{\chi}(\tau)$ can be approximated by
\begin{align*}
    \boldsymbol{\chi}(\tau)^{[3]} = \mathbf{dexp}_{\mathbf{X}^{[3]}(\tau) }\mathbf{X}'^{[3]}(\tau).
\end{align*}
Restricting this formulation to a Kirchhoff rod, the approximation of the rotation and the angular deformation fields become respectively
\begin{align}
    \mathbf{x}^{[3]}(\tau) &= (3\tau^2 - 2\tau^3)\mathbf{x}_1 + \tau(\tau - 1)^2 \boldsymbol{\kappa}_0 \notag \\ &+ (\tau^3 - \tau^2) \mathbf{dexp}^{-1}_{-\mathbf{x}_1} \boldsymbol{\kappa}(1) \label{eq:kirchhoff-interpolated-x} \\
    \boldsymbol{\kappa}^{[3]}(\tau) &= \mathbf{dexp}_{\mathbf{x}^{[3]}(\tau) }\mathbf{x}'^{[3]}(\tau).
    \label{eq:kirchhoff-interpolated-k}
    % \label{eq:kirchhoff-interpolated}
\end{align}
As for the exact model, $\mathbf{r}^{[3]}(\tau)$ can be integrated separately with Gaussian quadrature of order $s$
\begin{align*}
    \mathbf{r}^{[3]}(\tau) = \frac{\tau}{2} \sum_{i = 1}^{s} \alpha_i \, \text{exp} 
    \, \mathbf{x}^{[3]}\, \bar{p}_i,
    % \label{eq:kirchhoff-interpolated-r}
\end{align*}
with $\alpha_i$ Legendre weights and $\bar{p}_i := \frac{1}{2}(1+p_i)\tau$ with $p_i$ Gauss points. Adopting this notation, the approximated potential energy of a Kirchhoff rod is computed with the Gauss-Legendre quadrature as well
\begin{align}
  V_{\boldsymbol{\kappa}}^{[3]}(\bar{\mathbf{x}}, \boldsymbol{\kappa}_0, \boldsymbol{\kappa}_1)
  &= \frac{1}{2} \sum_{i=1}^{s} \alpha_i
       \bar{V}_{\boldsymbol{\kappa}}(\bar{p}_i, \mathbf{x}_1, \boldsymbol{\kappa}_0, \boldsymbol{\kappa}_1),
   \label{eq:pot-energy-interp-K}
\end{align}
with energy density
\begin{align*}
    \bar{V}_{\boldsymbol{\kappa}}(\boldsymbol{\kappa}^{[3]}(\tau)) = \frac{1}{2} (\boldsymbol{\kappa}^{[3]}(\tau) - \bar{\boldsymbol{\kappa}}^{[3]}(\tau))^{\text{T}} \mathbf{K}_{\boldsymbol{\kappa}} (\boldsymbol{\kappa}^{[3]}(\tau) - \bar{\boldsymbol{\kappa}}^{[3]}(\tau)).
\end{align*}
Finally, observe that this formulation yields to an analytical closed-form of the model~\cite{Mueller2024}. 

When considering manipulation tasks, where the rod is constrained at its extremities, the shape reconstruction problem can be formulated as a Boundary Value Problem (\textit{BVP}).
Given the terminal poses $\mathbf{H}_0 = \mathbf{I}$ and $\mathbf{H}_1$, the \textit{BVP} for the reduced Kirchhoff is satisfied when the geometric constraint 
\begin{align}
    \mathbf{g}(\boldsymbol{\kappa}_0, \boldsymbol{\kappa}_1) := \mathbf{r}^{[3]}(1) - \mathbf{r}_1 = \mathbf{0},
    \label{eq:geometric-constraint}
\end{align}
holds. Notice that the constraint involves only the end tip position, as orientation requirements are automatically satisfied through the interpolation structure. Therefore, the shape reconstruction problem reduces to finding the curvature values at the rod extremities $\boldsymbol{\kappa}(0)$ and $\boldsymbol{\kappa}(1)$, that satisfy the boundary tip position $\mathbf{r}_1$. Because multiple third-degree polynomial curves can satisfy the geometric constraint, the solution (i.e. shape) is selected as the minimum energy curve that satisfies the \textit{BVP}. 

The constrained energy minimization problem can be then formulated as
\begin{align}
    &\min_{\boldsymbol{\kappa}_0, \boldsymbol{\kappa}_1} \, V_{\boldsymbol{\kappa}}^{[3]}(\mathbf{x}_1, \boldsymbol{\kappa}_0, \boldsymbol{\kappa}_1) \quad \\
    & \, \,\text{s.t.} \quad \mathbf{g}(\boldsymbol{\kappa}_0, \boldsymbol{\kappa}_1) = \mathbf{0},
    \label{eq:minimization-problem}
\end{align}
where the potential energy $ V_{\boldsymbol{\kappa}}^{[3]}$ is obtained via quadrature as in Eq.~\eqref{eq:pot-energy-interp-K} and the geometric constraint $\mathbf{g}$ is defined in Eq.~\eqref{eq:geometric-constraint}. The associated Lagrangian function is
\begin{align*}
    \mathcal{L}(\boldsymbol{\kappa}_0, \boldsymbol{\kappa}_1,\boldsymbol{\lambda}) = V^{[3]}(\mathbf{x}_1, \boldsymbol{\kappa}_0, \boldsymbol{\kappa}_1) + \boldsymbol{\lambda}^{\text{T}}  \mathbf{g}(\boldsymbol{\kappa}_0, \boldsymbol{\kappa}_1),
\end{align*}
with corresponding stationarity conditions
\begin{align*}
    \begin{cases}
        \partial_{\boldsymbol{\kappa}_0} \mathcal{L}(\boldsymbol{\kappa}_0, \boldsymbol{\kappa}_1, \boldsymbol{\lambda}) &= \mathbf{0} \\
        \partial_{\boldsymbol{\kappa}_1} \mathcal{L}(\boldsymbol{\kappa}_0, \boldsymbol{\kappa}_1, \boldsymbol{\lambda}) &= \mathbf{0} \\
        \partial_{\boldsymbol{\lambda}} \mathcal{L}(\boldsymbol{\kappa}_0, \boldsymbol{\kappa}_1, \boldsymbol{\lambda}) &= \mathbf{0} .\\
        % \mathbf{J} \, (\boldsymbol{\kappa}_0 \, \boldsymbol{\kappa}_1 )^{\text{T}} &= \mathbf{0}.
    \end{cases}
\end{align*}
The minimization problem can be solved adopting a Newton-Raphson iterative scheme, updating the curvature parameters $\boldsymbol{\kappa}_0$ and $\boldsymbol{\kappa}_1$ until the geometric constraint in Eq.~\eqref{eq:geometric-constraint} is satisfied.

% \begin{remark}
%     The formulation hereby presented neglects the presence of external wrenches acting on the rod.
% \end{remark}

\subsection{Geometric Variable Strain Model}
\label{sec:GVS}
This section reviews the \textit{GVS} formulation~\cite{Boyer2019IEEE} using the notation of Table~\ref{table:list-of-symbols} to shed light on the main differences between the two models.
The kinematics and statics ODEs are
\begin{align}
    \text{\textit{Forward}:} \hspace{1.5cm} &\mathbf{H}'(\tau) = \mathbf{H}(\tau) \, \hat{\boldsymbol{\chi}}(\tau), \\ \\
    \text{\textit{Backward}:} \hspace{1.5cm} &\boldsymbol{\Lambda}(\tau) = \mathbf{K}(\tau) (\boldsymbol{\chi}(\tau) - \bar{\boldsymbol{\chi}}(\tau)) \\ 
    & \mathbf{Q}(\tau) = - \boldsymbol{\Phi}^{\text{T}} \mathbf{B}^{\text{T}} \boldsymbol{\Lambda}(\tau),
    \label{eq:GVS-fw-bw-integration}
\end{align}
that are not integrated concurrently, but forward and backward integrated one after the other. The deformation is approximated using the Ritz method by
\begin{align}
    \boldsymbol{\chi}(\tau) = \mathbf{B} \, \boldsymbol{\Phi}(\tau) \, \mathbf{X}(\tau),
    \label{eq:GVS-deformation-interp}
\end{align}
with $\mathbf{B}$ representing the enabled deformation modes, $\mathbf{X} \in \mathbb{R}^{n\times1}$ the vector of generalized coordinates and $\boldsymbol{\Phi}$ the matrix of shape functions
\begin{align*}
     \boldsymbol{\Phi} = \begin{pmatrix}
        P_n(\tau) & \mathbf{0}_{1 \times n \,(d - \, 1)} \\
        \hspace{2cm} \ddots \\
        \mathbf{0}_{1 \times  n \,(d - \, 1)} &  P_n(\tau)
     \end{pmatrix},
\end{align*}
with $d$ number of rod modes and $P_n$ polynomial basis of degree $n$.  
Notice that one can use a different degree in each row and a different polynomial basis. This latter can lead to different results, according to their numerical stability~\cite{Mathew2024}. 
This model is then solved for the boundary conditions
\begin{align}
    \begin{cases}
        & \mathbf{r}(\tau = 1) - \mathbf{r}_{\text{des}} = \mathbf{0} \\
        & (\mathbf{R}_{\text{des}}^{\text{T}} \, \mathbf{R}(\tau = 1) -  \mathbf{R}(\tau = 1)^{\text{T}} \, \mathbf{R}_{\text{des}} )^{\vee} = \mathbf{0}\\
        & \mathbf{K}_{\epsilon\epsilon} \, \mathbf{X} - \mathbf{Q}(\tau = 0) \, + \, \mathbf{J}(\tau=1)^{\text{T}}\boldsymbol{\Lambda}(\tau = 1) = \mathbf{0}.
    \end{cases}
    \label{eq:GVS-BCs}
\end{align}
Here $\mathbf{K}_{\epsilon\epsilon} = \int_{0}^{1} \boldsymbol{\Phi}^{\text{T}} \, \mathbf{B}^{\text{T}} \, \mathbf{K} \, \mathbf{B} \, \boldsymbol{\Phi} \, \text{d}\tau$ stands for the generalized stiffness matrix and $\boldsymbol{\Lambda}(\tau = 1)$ is the wrench imposing the constraint tip pose, expressed in terms of generalized coordinates through the Jacobian operator
\begin{align*}
    \mathbf{J}(\tau=1) = \text{Ad}_{\mathbf{H}(\tau=1)}^{-1} \int_{0}^{1}  \text{Ad}_{\mathbf{H}(\tau)} \, \mathbf{B} \, \boldsymbol{\Phi} \, \text{d}\tau.
\end{align*}

\begin{remark}
The strain-interpolated approach, recalled in Section~\ref{sec:interp-model}, approximates the canonical coordinates (Eq.~\ref{eq:kirchhoff-interpolated-x}) to retrieve the deformation (Eq.~\ref{eq:kirchhoff-interpolated-k}). Here, the \textit{BVP} is structured as a constrained energy minimization problem subject to the sole position boundary condition (Eq.~\ref{eq:minimization-problem}). Notice that solutions are found for initial and final curvature values that satisfy the \textit{BCs} and minimize the rod energy. The orientation is implicitly satisfied by the interpolation structure and the formulation hereby presented neglects the external loads.
Instead, the \textit{GVS} model, reviewed in Section~\ref{sec:GVS}, adopts the Ritz method to approximate deformations as linear combination of the shape functions and generalized coordinates (Eq.~\ref{eq:GVS-deformation-interp}). The \textit{BVP} is formulated as a root-finding problem subject to pose and equilibrium \textit{BCs} (Eq.~\ref{eq:GVS-BCs}), where the equations (Eq.~\ref{eq:GVS-fw-bw-integration}) are forward and backward integrated to balance external wrenches. Solutions are found for vectors of generalized coordinates and constraint wrenches that satisfy the \textit{BCs}. This model formulation allows interpolation degree and shape function modifications, leading to approximations of arbitrary order.

\end{remark}

\section{Evaluation}
\label{sec:evaluation-metrics}
When it comes to evaluate the performance of strain-based models for continuum robots, with a focus on manipulation, literature lacks of established benchmarks. Beside the flying rod proposed in~\cite{Simo1988_flyingbeam} and later reproduced in~\cite{Cesarek2013, Boyer2019IEEE}, no benchmark has been systematically adopted neither in simulation nor experimentally. However, this test is not suitable for the dual-manipulation scenarios considered in this work, where the rod is constrained at both ends. Consequently, to assess the performances of approximated inextensible Kirchhoff rod models, dedicated tests should give evidence to the allowed deformation modes. To this end, this work employs a set of quasi-static trajectories that first isolates individual deformation modes, and subsequently combines them. 

\subsection{Accuracy Metric} % \subsection{Evaluation Metrics}
Establishing a metric to retrieve performance insights on the models' behaviour requires defining a ground-truth.
In simulation, the reference is given by numerical exact solutions obtained solving the \textit{ODE} system in Equations~\eqref{eq: chi = H^-1 H'},~\eqref{eq:cosserat-constitutive},~\eqref{eq:cosserat-euler-equation}, here reported,
\begin{align*}
\begin{cases}
    &\mathbf{H}'(\tau) = \mathbf{H}(\tau) \, \hat{\boldsymbol{\chi}}(\tau) \\  & \boldsymbol{\Lambda}(\tau) = \mathbf{K}(\tau) (\boldsymbol{\chi}(\tau) - \bar{\boldsymbol{\chi}}(\tau)) \\
    &\boldsymbol{\Lambda}'(\tau) =  \operatorname{ad}_{\boldsymbol{\chi}(\tau)}^{\text{T}} {\boldsymbol{\Lambda}(\tau)} + \mathbf{W}(\tau),
\end{cases}
\end{align*}
subject to the boundary conditions
\begin{align*}
    \begin{cases}
        & \mathbf{r}(1) - \mathbf{r}_{\text{des}} = \mathbf{0} \\
        & (\mathbf{R}_{\text{des}}^{\text{T}} \, \mathbf{R}(1) -  \mathbf{R}(1)^{\text{T}} \, \mathbf{R}_{\text{des}} )^{\vee} = \mathbf{0},
    \end{cases}
\end{align*}
yielding a continuous curve describing the rod shape.

While typically accuracy is intended only in terms of one tip position, here tip poses constitute the boundary conditions, hence implicitly satisfied by the model solutions. Therefore, accuracy is evaluated according to cross-sectional pose error as 
\begin{align*}
    &e_{r,m}(\tau) = || \, \mathbf{r}_m (\tau) - \mathbf{r}_e (\tau) \, ||\, , \\
    &e_{x,m}(\tau) = \frac{1}{2} \operatorname{tr}(\mathbf{I} - \mathbf{R}_e^{\text{T}}(\tau)\mathbf{R}_m(\tau)) \, ,
\end{align*} 
with $e_{r,m}, \, e_{x,m}$ indicating cross-sectional position and orientation errors, respectively, and $\tau \in [0,1]$. Notice that quantities indicated with the subscript $(\cdot)_m$ are related to models, while the subscript $(\cdot)_e$ refers to the reference, being the exact model solutions in simulations and experimental measures, when coming to the experiments.
Adopting this metric, position and orientation errors can be graphically represented as curves varying over $\tau$. Despite this visualization being intuitive for analysing a single configuration, it is not suitable for quantitative comparison of several tests. Therefore, each configuration is also associated with the integral of the error,
\begin{align*}
    e_{r,m, \, \int} = \int_0^1 e_{r,m}(\tau) \, \text{d}\tau, \\
    e_{x,m, \, \int} = \int_0^1 e_{x,m}(\tau) \, \text{d}\tau,
\end{align*}
yielding a scalar measure of the total configuration error. Additionally, the maximum configuration errors for both position and orientation will be used, namely $e_{r,m, \, \text{max}}, \, e_{x,m, \, \text{max}}$.

While in simulation the number of comparison points can be arbitrarily selected, in experiments the number of comparison points is limited by the employed set-up and subjected to uncertainty. Therefore, the exact cross-sectional position $\mathbf{r}_e(\tau)$ is replaced by discrete measurement points $\mathbf{p}_e$, and the total configuration error is reformulated from an integral into a sum,
\begin{align}
    &e_{r,m}(i) = || \, \mathbf{r}_m (\tau) - \mathbf{p}_e (i) \, ||\, , \\
    &e_{r,m, \, \sum} = \sum_{i=1}^N \, e_{r,m}(N).
    \label{eq:error-exp}
\end{align}
Notice that cross-sectional orientation information is not available for the experiments, as commonly occurs in literature~\cite{FergusonRuckerWebster2024}. 

\subsection{Computational Time}
Evaluating the computational time provides insights on the models' suitability for real-time tasks. However, time analysis is strongly dependent on both software implementation and hardware features, making it difficult to obtain comparable results across different works. Indeed, to the best of authors' knowledge, the computational time analysis of approximated rod models does not foresee standardized tests expecting specific time results.
Therefore, even if a time comparison will be reported, we stress that this work primarily forces on providing a methodology to compare models on the accuracy-side first, rather than focusing on the computational aspects.

All presented models are implemented in MATLAB, with root-finding and minimization procedures handled by the software. A common numerical convergence threshold $10^{-7}$ is adopted across the models to ensure, as much as possible, a fair comparison.
In particular, the computational time is intended as the time required to find a solution to the \textit{BVP}, then it is not referred to a single iteration, while providing a warm start\footnote{On a MacBook Air 2025 with Apple M4 processor and 16 GBs RAM.}.

\section{Simulations} \label{sec:simulations}

This section presents a set of numerical simulations designed to evaluate the behaviour of the interpolated model. All simulations are performed in quasi-static conditions and in the absence of gravity, to focus on the rod deformation response. The proposed test trajectories are designed to selectively stress specific deformation modes, analysing models’ ability to capture bending, torsion, and combined deformation behaviours. To provide further comparison and performance evaluation, the predictions of the \textit{GVS} models are also included.

\subsection{Scenarios}
The simulations consider a fibreglass rod with circular cross-section, diameter $d = 2$~mm, length $L = 1$~m, Young's Modulus $E = 36.5$~GPa and shear Modulus $G = 3$~GPa. The \textit{GVS} model is tested with a monomial basis of third degree for both bending and torsion.

\subsubsection{Bending Simulations} \label{sec:simulations-bending}
Since the rod has a circular cross-section and the material is assumed isotropic, the two bending behaviours are supposed to be mechanically equivalent. Consequently, provided that the same number of degrees of freedom is used to approximate both bending curvatures, it is sufficient to test one bending direction only.
\begin{figure*}
    \centering
    \begin{minipage}{0.48\linewidth}
        \centering
        \includegraphics[width=\linewidth]{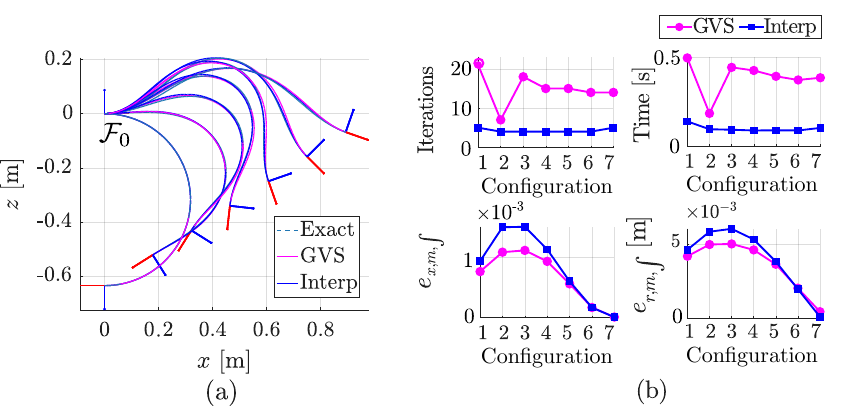}
        \caption{Bending test: rod configurations (a) and evaluation metrics plots (b). The \textit{GVS} is solved with third degree monomial basis. Although selecting a second degree monomial basis reduces computational time, it remains higher than the interpolated one. On the other hand, accuracy decreases to the same range of the interpolated.}
        \label{fig:1-curvatures}
    \end{minipage}
    \hfill
    \begin{minipage}{0.48\linewidth}
        \centering
        \includegraphics[width=0.95\linewidth]{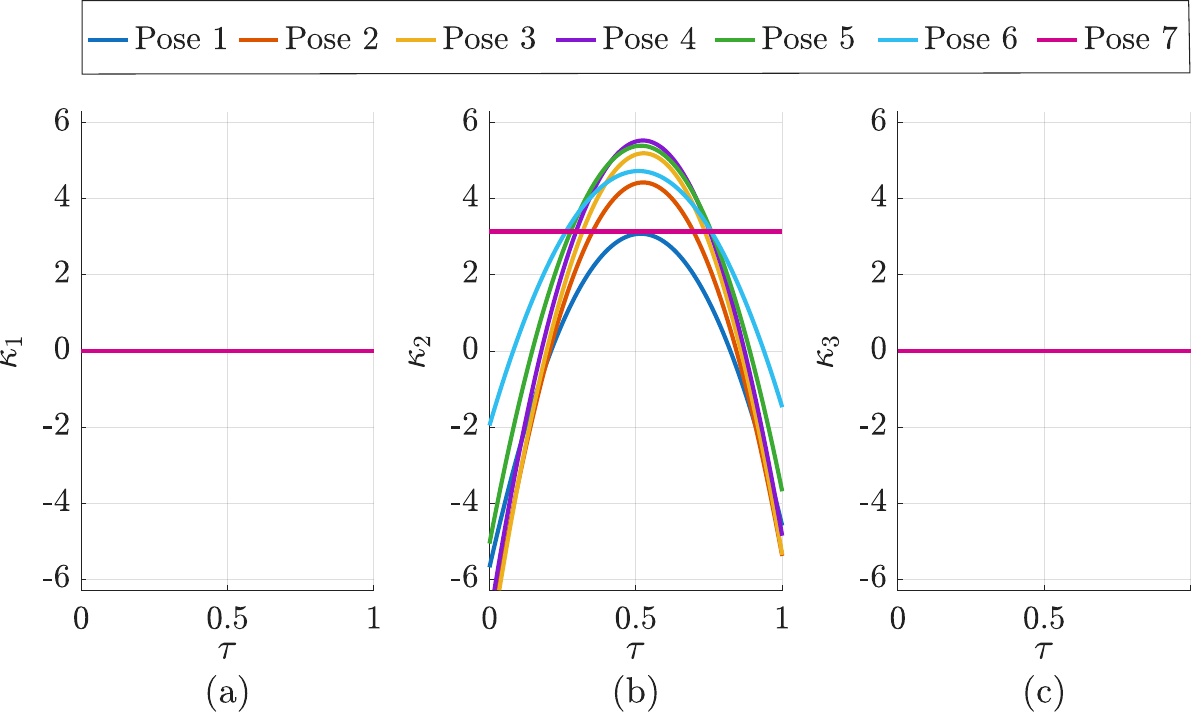}
        \caption{Values of $\kappa_1$ (a), $\kappa_2$ (b) and $\kappa_3$ (c) at each $\tau$ for the interpolated model.}
        \label{fig:kappa-values-1-curvature}
    \end{minipage}
\end{figure*}
Figure~\ref{fig:1-curvatures} shows a planar trajectory that excites the curvature $\kappa_2$ corresponding to bending about the $y-$axis. Figure~\ref{fig:1-curvatures}(a) depicts 7 rod configurations along the trajectory. Notice that the final one corresponds to the uniform bending configuration, with the rod lying on a circle of radius $1/\pi$ and exhibiting constant bending curvature $\kappa_2 = \pi$. 
Figure~\ref{fig:1-curvatures}(b), instead, reports the performance in terms of iteration number, computational time, integrated position and orientation errors.

The third configuration, which presents a high variation of curvature, is particularly challenging thus resulting in the maximum error for this trajectory. Specifically, this configuration yields a maximum position error of $1.64\%$ of the total rod length for the interpolated model and $1.03\%$ for the \textit{GVS} model, with mean total errors across the trajectory of $0.40\%$ and $0.36\%$, respectively.
In this scenario, the interpolated model proves to be a more efficient alternative, requiring on average $0.10$s and $4$ iterations, compared to $0.35$s and $15$ iterations, for the \textit{GVS} model. Reducing the order of the polynomial basis of the \textit{GVS} reduces the computational time, and the accuracy decreases at about the same value of the interpolated one. Instead, using Legendre polynomial of forth degree provides an even more accurate solution, within the same computational time range of the monomial basis.
Additionally, Figure~\ref{fig:kappa-values-1-curvature} reports the deformation values, \textit{i.e.} $\kappa_1$ (torsion), $\kappa_2$ (bending about the $y$-axis), and $\kappa_3$ (bending about the $z$-axis) for each snapshot along the trajectory, coherently showing that $\kappa_2$ becomes constant in the final configuration.

\begin{figure}[t]
    \centering
    \includegraphics[width=\columnwidth]{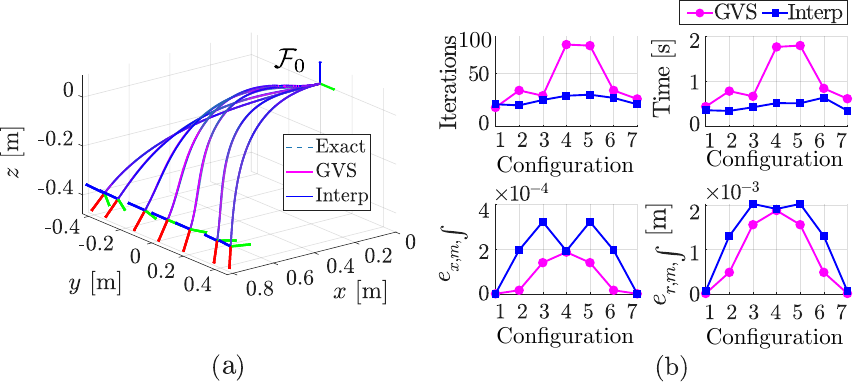}
    \caption{Two bending test: rod configurations (a) and evaluation metrics plots (b). The \textit{GVS} model is solved with thirds degree monomial basis.}
    \label{fig:2-curvatures}
    
\end{figure}
Increasing the number of stressed deformation modes, Figure~\ref{fig:2-curvatures}(a) illustrates a trajectory involving both bending curvatures. Similarly to the previous simulation, the interpolated model exhibits larger maximum position error $0.43\%$ compared to the \textit{GVS} model $0.35\%$ and mean total error of $0.12\%$ versus $0.08\%$. The maximum orientation errors remain $0.05\%$ for both models. Nevertheless, the interpolated computational efficiency remains notable, requiring on average $0.45$s and $28$ iterations against $0.99$s and $42$ iterations needed by the \textit{GVS} model. Notice that both time and errors increase with respect to one bending only.  

\subsubsection{Torsion Simulation}
The remaining deformation mode to be tested is the torsional curvature $\kappa_1$. In this scenario, the rod is subjected to an increasing twist at the tip, starting from the undeformed configuration. It is worth noticing that a big twisting angle would require axial elongation, which is not captured by the inextensible Kirchhoff rod model considered in this work. 
However, to test models, the simulation is carried out imposing a final twist of $\pi$ at the tip. Both models show a high level of agreement with the exact solution, with mean and maximum errors below $0.01\%$ of the total rod length. As for previous simulations, the interpolated model requires less time for convergence, with an average time of $0.14$s versus $0.25$s of the \textit{GVS} model. 

\subsubsection{Bending-torsion Simulation}
\begin{figure*}
    \centering
    \includegraphics[width=\linewidth]{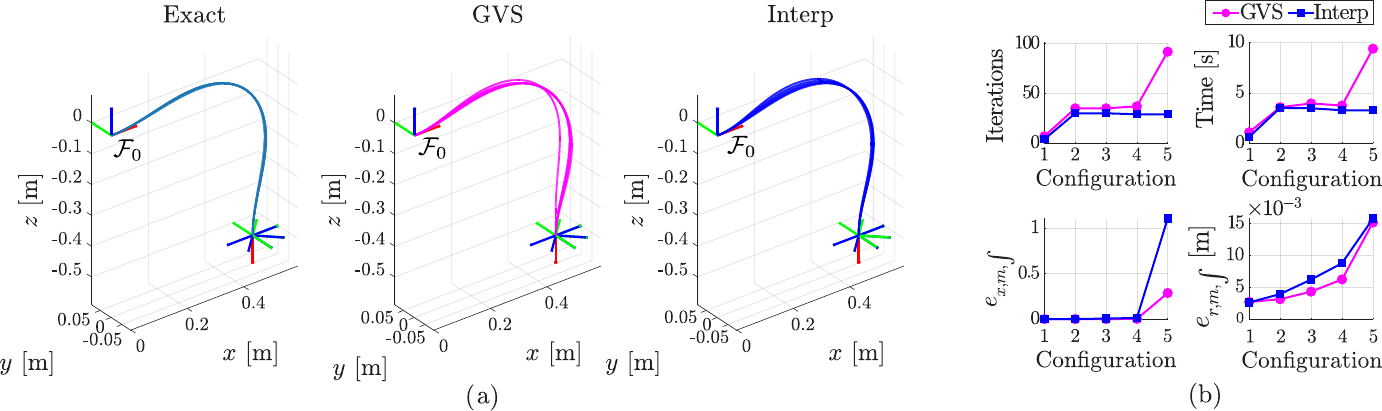}
    \caption{Bending-torsion test: rod configurations for the exact model (a-left), \textit{GVS} model (a-middle) and interpolated model (a-left), together with evaluation metrics (b). The \textit{GVS} model is solved with third degree Legendre polynomial basis.}
    \label{fig:bending-torsion}
\end{figure*}

Combined deformation effects are common in rod manipulation tasks and are responsible for most of the geometric non-linearities in statics. For this reason, assessing the ability of the models to handle multiple deformation modes simultaneously is of particular interest. A benchmark addressing the coupling between torsion and bending is presented in \cite{Boyer2019IEEE, Boyer2003}, where a rod is twisted and its ends are subsequently pushed toward each other. In that case, an extensible rod model is adopted. However, since this work considers inextensible Kirchhoff rods, a different coupled bending-torsion test, that does not involve compression, is shown in Figure~\ref{fig:bending-torsion}(a). One rod tip is kept fixed, while the other is initially displaced to $\mathbf{r}_{\text{des}} = L\,\begin{pmatrix}0.5 & 0 & 0.5\end{pmatrix}^{\text{T}}$, $\mathbf{x}_{\text{des}} = \begin{pmatrix}0  & \pi/2 & 0 \end{pmatrix}^{\text{T}}$ and subsequently twisted in increments $\theta_{step} = \pi/4$ until a final rotation of $\theta = \pi$ is reached.
In this simulation, the \textit{GVS} monomial basis was changed to Legendre polynomials with three \textit{DOF} for every deformation, to carry out the simulation.
Figure~\ref{fig:bending-torsion}(b) shows the results.
The interpolated model requires less computational time and iterations, on average $2.82$s and $24$ iterations w.r.t. $4.35$s and $41$ iterations of the \textit{GVS}, where for the final configuration $\theta = \pi$ it takes almost three times longer than the interpolated. For both models, accuracy decreases as torsion increases, with smaller errors associated with the \textit{GVS} predictions with maximum position error of $2.30\%$, versus $3.01\%$ for the interpolated, and mean error $0.62\%$, versus $0.75\%$. 

%%%%%%%%%%%%%%%%%%%%%%%
\begin{table*}[]
\centering
\caption{Simulation results Summary.
Errors are expressed as percentage of the total rod length $L$.}
\label{tab:simulation-recap}
% \begin{tabular}{lcccccc}
\begin{tabularx}{\textwidth}{l 
                                 >{\centering\arraybackslash}X 
                                 >{\centering\arraybackslash}X 
                                 >{\centering\arraybackslash}X
                                 >{\centering\arraybackslash}X 
                                 >{\centering\arraybackslash}X 
                                 >{\centering\arraybackslash}X}
\toprule
TEST CASE 
& \multicolumn{2}{c}{MAX POSITION ERR [\%]} 
& \multicolumn{2}{c}{MEAN POSITION ERR [\%]} 
& \multicolumn{2}{c}{COMPUTATION} \\
\cmidrule(lr){2-3} \cmidrule(lr){4-5} \cmidrule(lr){6-7}
 & Interp. & GVS & Interp. & GVS & Time [s] & Iter. \\
\midrule
Bending ($\kappa_2$) 
& 1.64 & 1.03 
& 0.40 & 0.35 
& 0.10 / 0.38 & 4 / 15 \\

Bending  ($\kappa_2,\kappa_3$) 
& 0.43 & 0.35 
& 0.12  & 0.08  
& 0.45 / 0.99 & 28 / 42  \\

Torsion ($\kappa_1$, $\theta \leq \pi$) 
& $<0.01$ & $<0.01$
& $<0.01$ & $<0.01$
& 0.14 / 0.25 & 5 / 9 \\

Bending-torsion ($\kappa_1,\kappa_2,\kappa_3$) 
&  3.01 & 2.30   
&  0.75 & 0.62  
&  2.82 / 4.35 & 24 / 41\\
\bottomrule
% \end{tabular}
\end{tabularx}
\end{table*}

\subsection{Discussion of Simulation Results}
The numerical simulations presented in this section, whose results are summarised in Table~\ref{tab:simulation-recap}, provide an assessment of the interpolated model performance across increasing levels of deformation complexity, with further comparison with the \textit{GVS} model.

In bending scenarios, both formulations exhibit comparable accuracy, while the interpolated model achieves lower computational time and fewer iterations. A similar behaviour is observed for pure torsion, where both models closely match the exact solution within the limits of the inextensibility assumption. In terms of time, combined effects require more computational effort with respect to bending alone and lead to degradation of accuracy for both approaches. In these cases, the \textit{GVS} model generally demonstrates higher accuracy, whereas the interpolated model maintains computational advantage. 
As the \textit{GVS} model allows selecting type and degree of the polynomial basis, different combinations were tested. In the single bending test, decreasing the monomial basis degree reduces computational time at the cost of accuracy, while in the bending-torsion test, Legendre polynomial basis are selected over the monomial ones to gain accuracy. However, among the tested basis, no choice represented at the same time a faster and more accurate alternative to the interpolated model.

Overall, these simulations demonstrate that the interpolated model meets the trade-off between accuracy and computational efficiency, with a mean total error below $0.75\%$ of the rod length compared to $0.62\%$ of the \textit{GVS} model and half of its computational time.  

\section{Experiments}
\label{sec:experiments}

The simulations are translated into experimental tests using a robotic arm to move one extremity of the rod, while capturing its shape using a vision-based system. In the following, the interpolated model and \textit{GVS} predictions are compared to the experimental measurements to asses their ability to capture real-world rod configurations.

\begin{table}[ht]
    \centering
    \small
    \renewcommand{\arraystretch}{1.1}
    \caption{Nitinol Rod parameters}
    \begin{tabularx}{\columnwidth}{>{\centering\arraybackslash}X 
                                   >{\centering\arraybackslash}X}
        \toprule
        PARAMETER & VALUE \\
        \midrule
        Rod Length& $0.955$ m \\
        Rod Diameter & $1.5\times10^{-3}$ m \\
        % Rod Weight & \textcolor{red}{???} g \\
        % Marker Weight & $1.70$ g \\
        Number of Markers & $7$ \\
        Distance Between Markers & $0.119$ m \\ 
        Young’s Modulus\footnotemark[\value{footnote}] $E$ & $80$ GPa \\
        Shear Modulus\footnotemark[\value{footnote}] $G$ & $30$ GPa \\
        % Second Moment of Area $I$ & $7.85 \times 10^{-13}$ m$^4$  \\
        \bottomrule
    \end{tabularx}
    \label{table:nitinol-params}
\end{table}
\begin{table}[ht]
    \centering
    \small
    \renewcommand{\arraystretch}{1.1}
    \caption{Fibreglass Rod parameters}
    \begin{tabularx}{\columnwidth}{>{\centering\arraybackslash}X 
                                   >{\centering\arraybackslash}X}
        \toprule
        PARAMETER & VALUE \\
        \midrule
        Rod Length& $0.885$ m \\
        Rod Diameter & $2\times10^{-3}$ m \\
        % Rod Weight & $6.25$ g \\
        % Marker Weight & $1.70$ g \\
        Number of Markers & $6$ \\
        Distance Between Markers & $0.126$ m \\ 
        Young’s Modulus\footnotemark $E$ & $36.5$ GPa \\
        Shear Modulus\footnotemark[\value{footnote}] $G$ & $3$ GPa \\
        % Second Moment of Area $I$ & $7.85 \times 10^{-13}$ m$^4$  \\
        \bottomrule
    \end{tabularx}
    \label{table:fibreglass-params}
\end{table}

\begin{figure}[t]
    \centering
    \includegraphics[scale=1]{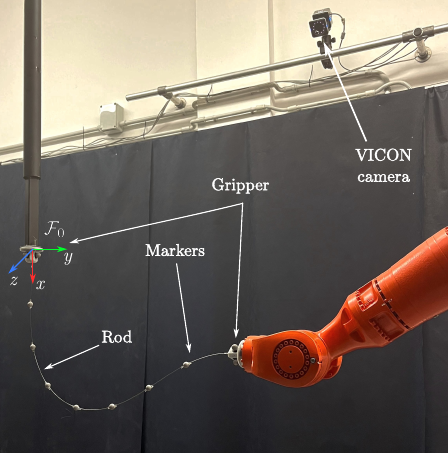}
    \caption{Experimental set-up: a nitinol rod is actuated by a KUKA robotic manipulator, while the 3D positions of reflective markers are captured by a multi-camera motion tracking system (VICON). Both clamps are 3D-printed in PLA and are designed with an ad-hoc geometry that matches the rod diameter, ensuring precise alignment and repeatable boundary conditions.}
    \label{fig:setup}
\end{figure}

\subsection{Set-up}
\label{sec:set-up}
To replicate dual manipulation tasks, we considered a rod with one end-tip fixed while a $6-$\textit{DOF} manipulator\footnote{KUKA manipulator KR 20 R3100.} moves the other to change the rod's shape.
Simultaneously, a VICON Vero v2.2 motion capture system with eight cameras records the positions of reflective markers placed along the rod. The system outputs the 3D positions of the markers along the rod, as well as the positions and orientations of the tips. The latter are reconstructed using three markers placed on the grippers.
The set-up is shown in Figure~\ref{fig:setup}

\footnotetext{Parameter estimated with three-point bending test on a Zwick Roell ProLine universal testing machine.}

\subsubsection{Markers}
In principle, increasing the number of markers provides finer information about the rod’s shape; however, one has to deal with practical constraints limiting their number. Firstly, markers must be sufficiently spaced for the cameras to distinguish them. This value was experimentally determined to be approximately $50$~mm. Secondly, each marker adds mass to the rod, potentially altering the expected shape. Therefore, since the experiments involving continuum elements typically do not embed concentrated masses, the marker arrangement is designed to minimize their influence, effectively allowing to neglect their weight. 
Indeed, the markers have a precise passing-through hole to be directly inserted in the rod. Moreover, small pieces of modelling clay are used to prevent any slippage during tests. It is clear that, even with careful placement, millimetric uncertainties in marker positions can affect the measurements and therefore the effectiveness of the comparison. Therefore, these uncertainties are minimized by recording the reference configuration (e.g. straight rod) to verify that the markers match their intended positions.

\subsubsection{Systematic Uncertainties}
Figure~\ref{fig:exp_ref} and Figure~\ref{fig:exp_cc} show the straight rod and the uniform bending (constant curvature) configurations.
Both are commonly used as first reference to verify a rod model. 
However, in this case, they are used to estimate and correct the uncertainties in the set-up to reduce their effect in the results.
The straight rod, Figure~\ref{fig:exp_ref}, shows errors in the range $2-5$~mm for both models ($0.21\%-0.53\%$ of the total length of the rod) primarily attributed to an angular misalignment of the rod support, causing an initial bending which would not be captured by the models. Analogously, the uniform bending configuration, Figure~\ref{fig:exp_cc}, shows errors in the range $1-7$~mm (below $0.75\%$ of the rod length) for both models.

\subsubsection{Rod}
Two different rods, in terms of geometry (length and diameter) and material, are used to broaden the experimental scenarios and understand whether the model performance depends on these parameters. Notice that $6$ markers were placed on a fibreglass rod while $7$ on the nitinol one, yielding a maximum deflection due to the marker weights below $1\%$ of the total length for each rod. Tables~\ref{table:nitinol-params} and~\ref{table:nitinol-params} reports the particular data relative to the two rods while Sections~\ref{sec:exp-nitinol} and~\ref{sec:exp-fibreglass} discuss their experiments.

\subsection{Experiments and Results with the Nitinol Rod} %Experimental Analysis}
\label{sec:exp-nitinol}
Starting with nitinol, in this section two quasi-static trajectories, both exciting multiple deformation modes, are illustrated.

Some experimental trajectories using the Nitinol rod (Table~\ref{table:nitinol-params}) are here presented, while results with the fibreglass rod will be discussed in the results Section~\ref{sec:exp-results}.
Firstly, two configuration tests are introduced to verify markers positions along the rod, and the error distribution in well-known configurations, for which analytical solutions are available. Subsequently, two quasi-static trajectories exciting multiple deformation modes are illustrated.
\begin{figure*}[h]
    \centering
    \begin{minipage}{0.48\linewidth}
        \centering
        \includegraphics[width=\linewidth]{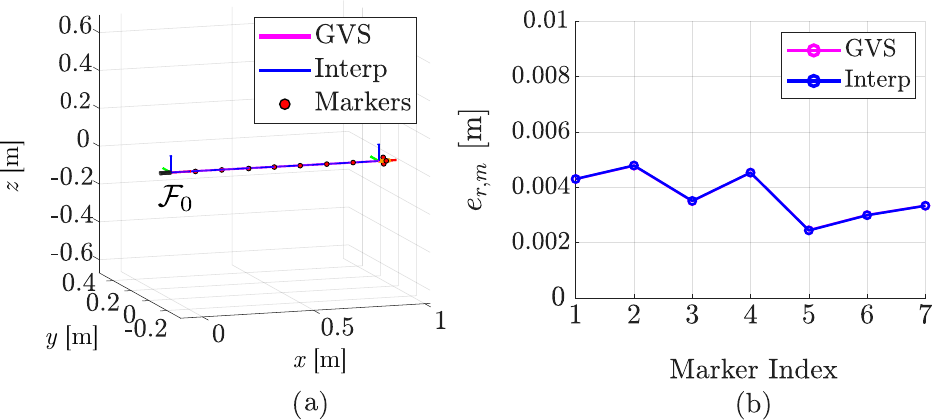}
        \caption{Straight rod experiment: configuration (a) and position error with respect to markers (b).}
        \label{fig:exp_ref}
    \end{minipage}
    \hfill
    \begin{minipage}{0.48\linewidth}
        \centering
        \includegraphics[width=\linewidth]{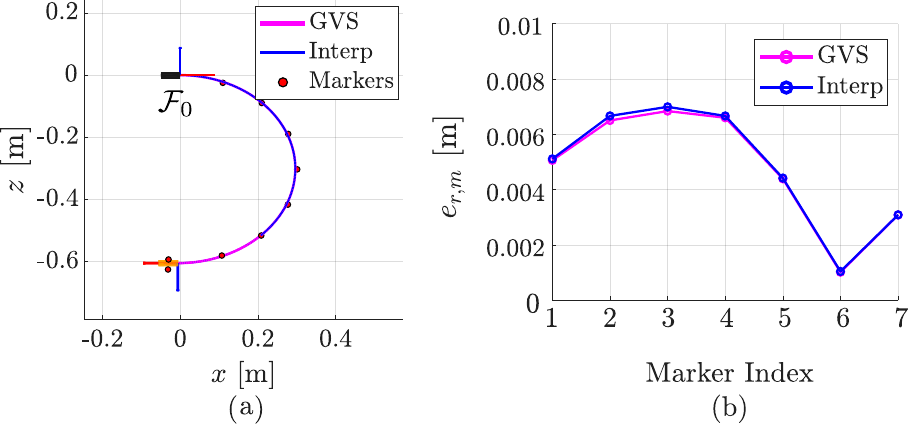}
        \caption{Uniform bending experiment: configuration (a),  position error with respect to markers (b).}
        \label{fig:exp_cc}
    \end{minipage}
\end{figure*}

\begin{figure*}[]
    \centering
    \includegraphics[width = \linewidth]{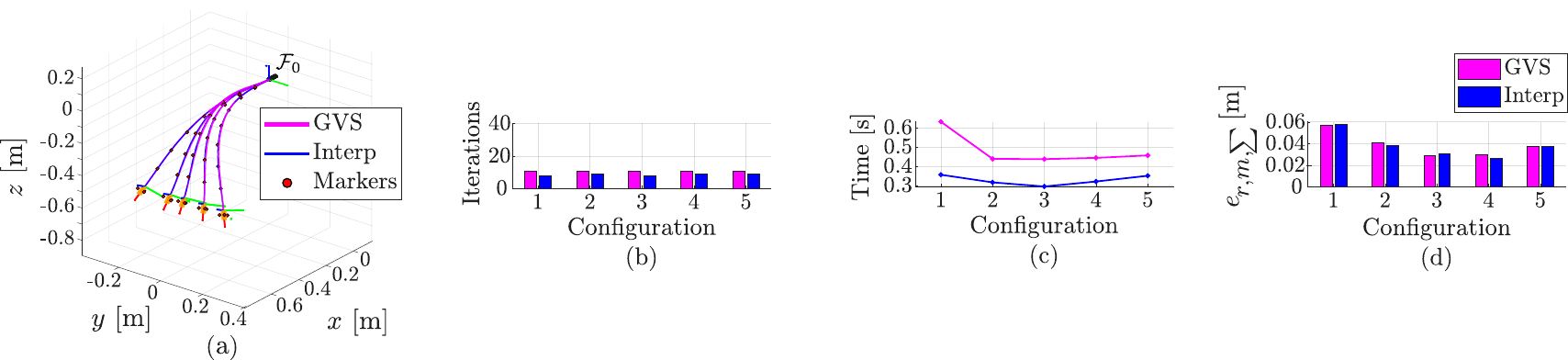}
    \caption{Two-bending experiment: configurations (a) and performances (b) in terms of iteration number, execution time and total configuration error. To be notice the symmetry of the results on figures (c) and (d) due to the imposed trajectory. The same happened above in Figure 4. The \textit{GVS} model is solved with second degree monomial basis.}
    \label{fig:exp_traj_2b}
\end{figure*}

\subsubsection{Two-bending Experiment}
Five snapshots of the bending trajectory simulated in Figure~\ref{fig:2-curvatures} are replicated experimentally and shown in Figure~\ref{fig:exp_traj_2b}(a). On the right side, the same figure provides the model performance, in terms of iterations Figure~\ref{fig:exp_traj_2b}(b), computation time Figure~\ref{fig:exp_traj_2b}(c) and total error Figure~\ref{fig:exp_traj_2b}(d), with colour legend coherent with the configuration plot.  
Overall, the mean error of the interpolated model is $0.58\%$ of the rod length, while the maximum position error is $1.39\%$. Analogously, the \textit{GVS} shows $0.59\%$ of mean error and maximum error of $1.38\%$ in the same configuration. In terms of computational performance, the interpolated model provides again a faster solution achieving a mean time of $0.32$s against $0.48$s for the \textit{GVS}.

\subsubsection{Bending-torsion Experiment}
\begin{figure*}
    \centering
    \includegraphics[width=\linewidth]{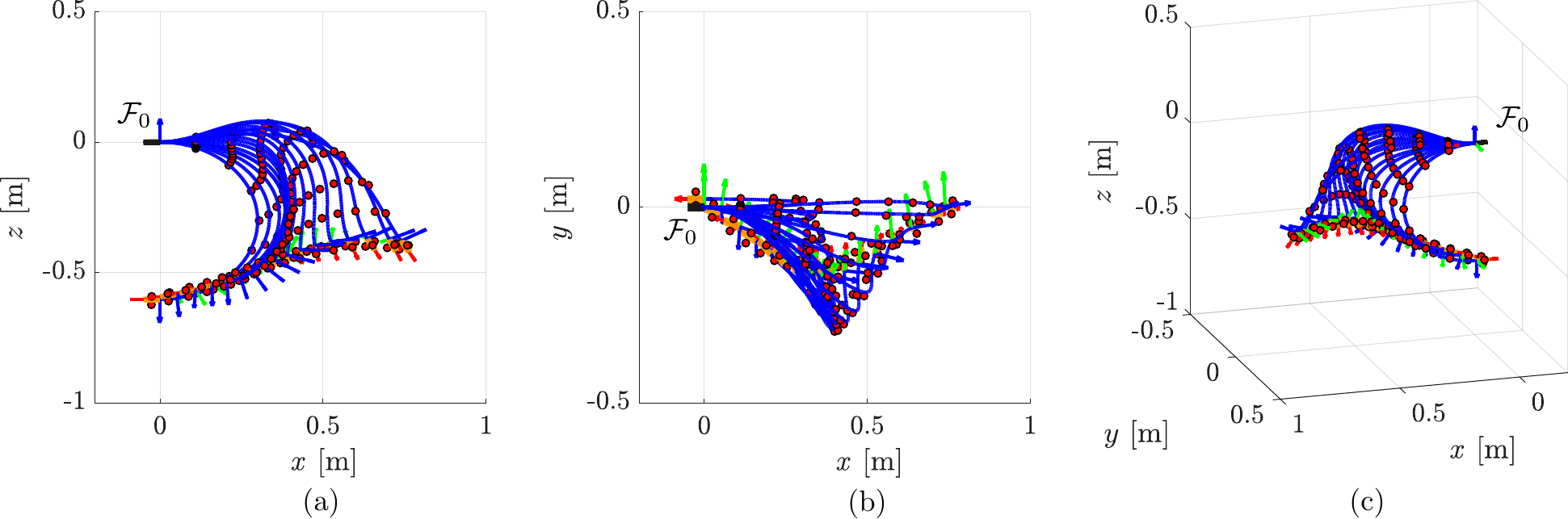}
    \caption{Snapshot of a trajectory inducing bending and torsional effects simultaneously. \\}
    \label{fig:exp_traj3}

    \includegraphics[width=\linewidth]{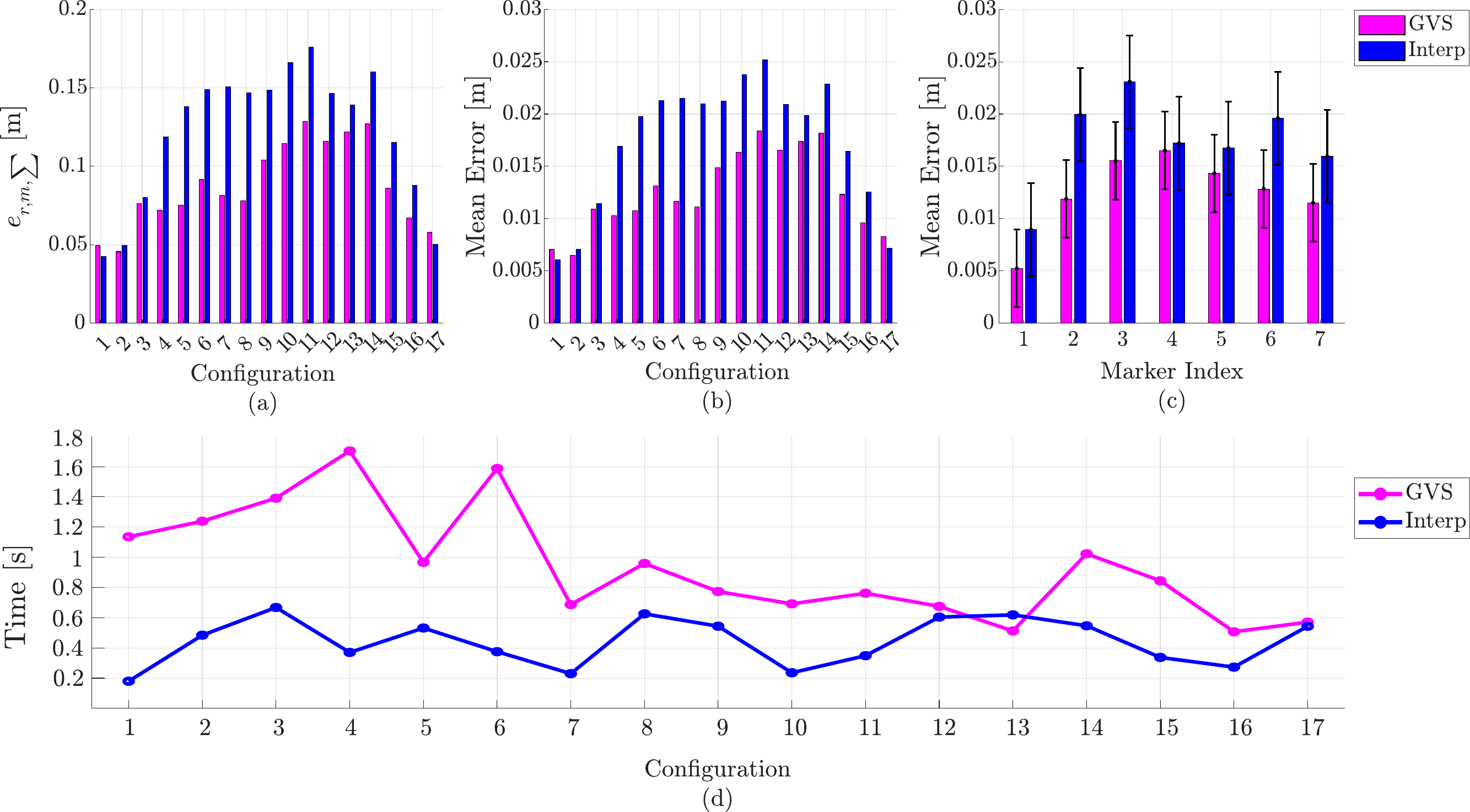}
    \caption{Bending-torsion experiment: configurations total error (a), configurations total error (b), mean markers error (c) and computational time (d). The \textit{GVS} model is solved with third degree monomial basis.}
    \label{fig:exp_traj3_error}
\end{figure*}

Figure~\ref{fig:exp_traj3} shows 17 snapshots of a trajectory that tests the coupling between bending and torsion from three different views.
The trajectory is obtained moving the robot from a generic rod configuration dominated by bending deformations to the uniform bending one. 
During the robot motion, the rod experiences bending about both axis and torsional effects. Indeed, from the two views of Figure~\ref{fig:exp_traj3}(a) and Figure~\ref{fig:exp_traj3}(b), it is possible to notice that the tip frame is twisted while bending.

Meanwhile, for each configuration Figure~\ref{fig:exp_traj3_error}(a) shows the total integral error and Figure~\ref{fig:exp_traj3_error}(b) the mean error. The first and last configurations are associated with lower error, mainly due to the reduced effect of the tip twist. Conversely, the middle configurations have pronounced torsion profiles, resulting in higher uncertainty from the models. Overall, the maximum and mean configuration errors are $3.62\%$ and $1.86\%$ for the interpolated and $3.03\%$ and $1.32\%$ for the \textit{GVS}. 
Figure~\ref{fig:exp_traj3_error}(c) reports the mean error and standard deviation for each of the seven markers placed on the rod, providing insight on the distribution error along the rod. Here, the marker 1 is the closest to the support, while the marker 7 is the closest to the robot.
The markers associated with higher uncertainty are the ones corresponding to the first half of the rod, confirming the previous results in Figure~\ref{fig:exp_ref} and~\ref{fig:exp_cc}. In terms of computational time, see Figure~\ref{fig:exp_traj3_error}(d), the interpolated model shows higher performance, with mean time of $0.64$s compared to $1.14$s of the \textit{GVS} model. Particularly, the configurations associated with higher torsional effects (configurations 12, 13 and 14) require more computational effort for the interpolated model.

\subsection{Experiments and Results with the Fibreglass Rod}
\label{sec:exp-fibreglass}
\begin{figure}[t]
    \centering
    \includegraphics[width=\columnwidth]{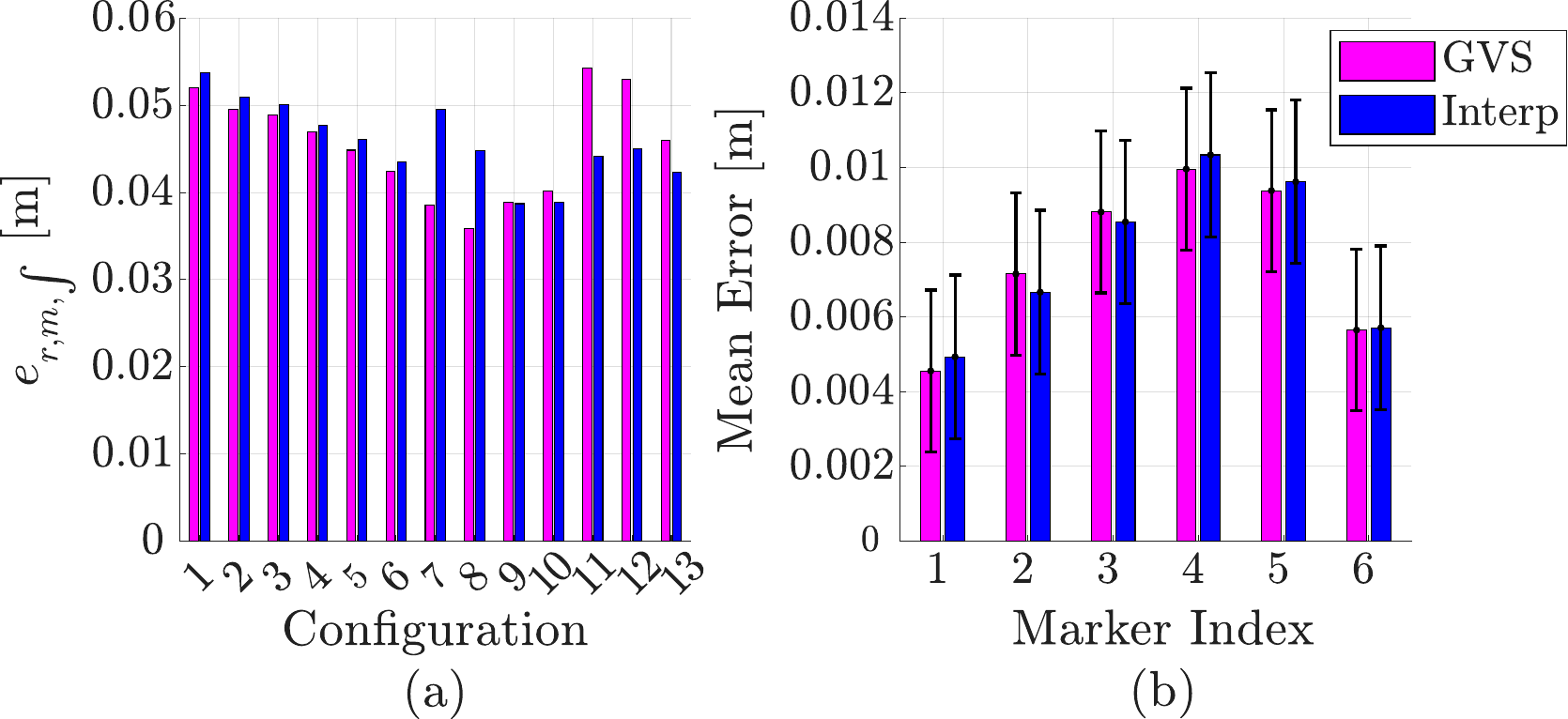}
    \caption{Fibreglass experiments: configurations total error (a) and markers mean error (b). The \textit{GVS} model solutions are found with second degree monomial basis.}
    \label{fig:exp_fibreglass}
\end{figure}

Other experiments were conducted using a fibreglass rod using six markers; see Table~\ref{table:fibreglass-params} for the specific parameters used in the experiments. Being the fibreglass more prone to breakage than the nitinol rod, the tested configurations involve primarily bending deformation rather than high torsional effects. Figure~\ref{fig:exp_fibreglass}(a) shows the results for 13 generic configurations. The interpolated model achieves maximum configuration error of $1.38\%$ of the rod length, while the \textit{GVS}, tested with second degree monomial basis, reaches $1.42\%$. The mean computation time is $0.34$s for the interpolated and $0.42$s for the \textit{GVS}. As for the other experiments, Figure~\ref{fig:exp_fibreglass}(b) confirms larger errors associated with the central portion of the rod, with the fourth marker being the one with higher error.

\begin{table*}[t]
\centering
\caption{Experiments Results Summary.
Errors are expressed as percentage of the total rod length $L$.}
\label{tab:exp-recap}
% \begin{tabular}{lcccccc}
\begin{tabularx}{\textwidth}{l 
                                 >{\centering\arraybackslash}X 
                                 >{\centering\arraybackslash}X 
                                 >{\centering\arraybackslash}X
                                 >{\centering\arraybackslash}X 
                                 >{\centering\arraybackslash}X 
                                 >{\centering\arraybackslash}X}
\toprule
TEST CASE 
& \multicolumn{2}{c}{MAX POSITION ERR [\%]} 
& \multicolumn{2}{c}{MEAN POSITION ERR [\%]} 
& \multicolumn{2}{c}{COMPUTATIONAL TIME [s]} \\
\cmidrule(lr){2-3} \cmidrule(lr){4-5} \cmidrule(lr){6-7}
 & Interp. & GVS & Interp. & GVS & Interp. & GVS \\
\midrule
Fibreglass Bending ($\kappa_2,\kappa_3$) 
& 1.38 & 1.42 
& 0.86  & 0.86  
& 0.34 & 0.42  \\

Nitinol Bending ($\kappa_2$, $\kappa_3$) 
& 1.39 & 1.38 
& 0.58 & 0.59 
& 0.32 & 0.48 \\

Nitinol Bending-torsion ($\kappa_1,\kappa_2,\kappa_3$) 
&  3.62 & 3.03   
&  1.86 & 1.32  
&  0.64 & 1.14\\

\bottomrule
% \end{tabular}
\end{tabularx}
\end{table*}

\subsection{Discussion of Experimental Results} \label{sec:exp-results}

The experimental results for both materials are summarized in Table~\ref{tab:exp-recap},
which reports computational time, maximum and mean errors. Note that errors are expressed as a percentage of the total rod length, and computational times are reported per configuration rather than per iteration. Furthermore, while the GVS model accounts for gravity, the interpolated model formulation does not incorporate external forces.

As shown in the first two rows of Table~\ref{tab:exp-recap}, the results remain coherent across the two materials, proving that the model's accuracy does not depend on the material choice. 
For fibreglass, the maximum errors for the interpolated and \textit{GVS} models are $1.38\%$ and $1.42\%$, with the same mean of $0.86\%$ while, for nitinol, the maximum errors are $1.39\%$ and $1.38\%$ with means $0.58\%$ and $0.59\%$, respectively.
The computational time is also comparable for the two materials, averaging $0.33$s for the interpolated model and $0.45$s for the \textit{GVS} model.
Instead, when torsion is present, convergence becomes more demanding due to the inextensibility assumption. As the models strive to find a solution, computational effort increases and accuracy degrades. Specifically, the maximum errors rise to the $3.62\%$ for the interpolated model and to the $3.03\%$ for the \textit{GVS} model, while average computational times increase to $0.64$s and $1.14$s, respectively.

Overall, the experimental trends align with simulation results, showing that both error and the computational cost grow with increased torsion. Consistently with the experiments, the \textit{GVS} model provides more accurate solutions, while the interpolated model converges faster. As expected, the simulation errors reported in Table~\ref{tab:simulation-recap} are lower than their experimental counterparts, with maximum error in bending simulation of $0.43\%$ for the interpolated and $0.35\%$ for the \textit{GVS} and maximum error in bending-torsion simulation of $3.01\%$ and $2.30\%$, respectively. 
The discrepancy between simulations and experiments, is not primarily due to gravity, which results in a minor error when comparing the two models, but it is rather due to the presence of unquantifiable uncertainties.

\section{Conclusion}
\label{sec:conclusion}

This work proposes an experimental methodology to evaluate strain-based models and adopts it for the performance assessment of the interpolated ones. The experimental framework uses a robotic arm to move one tip of a rod in space, while the other is clamped, thus replicating dual-arm manipulation tasks. A motion capture system is used to retrieve the rod shape through reflective markers. This approach allows direct measurement of the configuration avoiding the use of expensive sensors to test the models. 
The predictions of the interpolated model show high adherence to the experimental data, with mean and maximum position errors of $0.58\%$ and $1.39\%$ of the rod length in bending dominated scenarios and $1.86\%$ and $3.62\%$ when torsional effects are significant, providing a range of applicability of this model. 
Further comparison with the Geometric-Variable Strain model is included in the analysis, showing comparable accuracy results, with mean $0.59\%$ and maximum $1.38\%$ for bending scenarios and $1.32\%$ and $3.03\%$ for torsional ones. 
The presence of torsional effects lowers the accuracy and computational performance of both models, trying to satisfy the inextensibility assumption.
Considering the computational time, the interpolated model proves to be an efficient alternative, to the \textit{GVS} model with $0.64$s compared to $1.14$s, with equal number of degrees of freedom.
Despite neglecting the presence of external forces, the interpolated model provides an accurate and efficient solution.
Leveraging the model’s computational efficiency, subsequent research will focus on the design of a visual servoing control for real-time dual manipulation tasks, possibly in presence of external disturbances.

\appendices
\section{Lie Group and Lie Algebra Properties} \label{app:Lie group}

This paper relies on several well-established relations in robotics. For completeness and clarity, this appendix gathers these relations and reformulates them within a unified notation framework.

An first important relation linking $se(3)$ and $SE(3)$ is given by the exponential map $\exp:se(3) \rightarrow SE(3)$, which expresses the rigid motion of a body in terms of twists
\begin{align*}
    \exp(\hat{\mathbf{X}}) &= \begin{pmatrix}
    \sum_{i=0}^{\infty} \frac{1}{i!}\tilde{\mathbf{x}}^{i} & \sum_{i=0}^{\infty} \frac{1}{(i+1)!}\tilde{\mathbf{x}}^{i}\mathbf{y} \\
    0 & 1
    \end{pmatrix} \\ &= \begin{pmatrix}
    \exp\tilde{\mathbf{x}} & \mathbf{dexp}_{\mathbf{x}}\mathbf{y} \\
    0 & 1
    \end{pmatrix},
\end{align*}
where $\mathbf{dexp}$ represent the differential of the exponential map
\begin{align*}
    \mathbf{dexp}_{\mathbf{x}} &= \sum_{i=0}^{\infty} \frac{1}{(i+1)!}\text{ad}_{\tilde{\mathbf{x}}}^{i} 
    \\ &= \begin{pmatrix}
    \mathbf{dexp}_{\mathbf{x}} & 0 \\
    (D_{\mathbf{x}}\mathbf{dexp})(\mathbf{y}) & \mathbf{dexp}_{\mathbf{x}}
    \end{pmatrix},
\end{align*}
which also admits the inverse
\begin{align*}
    \mathbf{dexp}_{\mathbf{x}}^{-1} = \begin{pmatrix}
    \mathbf{dexp}_{\mathbf{x}}^{-1} & 0 \\
    (D_{\mathbf{x}}\mathbf{dexp}^{-1})(\mathbf{y}) & \mathbf{dexp}_{\mathbf{x}}^{-1}
    \end{pmatrix}.
\end{align*}
Two equivalent mappings for $\exp$ and $\mathbf{dexp}$, associated to $so(3)$ and $SO(3)$ are defined in closed form in~\cite{Mueller2021}. 

The Lie algebra adjoint operator $\mathrm{ad}: se(3)\times se(3)\rightarrow se(3)$
is defined via the Lie bracket
\begin{align*}
    \mathrm{ad}_{\hat{\mathbf{X}}}(\hat{\mathbf{Y}})
    := [\hat{\mathbf{X}},\hat{\mathbf{Y}}]
    = \hat{\mathbf{X}}\hat{\mathbf{Y}} - \hat{\mathbf{Y}}\hat{\mathbf{X}} .
\end{align*}
In vector representation, for $\mathbf{X}=(\mathbf{x},\mathbf{y})\in\mathbb{R}^6$,
the corresponding matrix form is
\begin{align*}
    \mathrm{ad}_{\mathbf{X}} =
    \begin{pmatrix}
        \tilde{\mathbf{x}} & 0 \\
        \tilde{\mathbf{y}} & \tilde{\mathbf{x}}
    \end{pmatrix}.
\end{align*}
For $\mathbf{H} \in SE(3)$, the adjoint representation matrix is instead defined as
\begin{align*}
    \mathrm{Ad}_{\mathbf{H}} =
    \begin{pmatrix}
        \mathbf{R} & 0 \\
        \tilde{\mathbf{r}}\mathbf{R} & \mathbf{R}
    \end{pmatrix}.
\end{align*}
Finally, the adjoint operators satisfy the identity
\begin{align*}
    \mathrm{Ad}_{\operatorname{exp}(\hat{\mathbf{X}})} = \operatorname{exp}(\operatorname{ad}_{\mathbf{X}}).
\end{align*}

\bibliographystyle{ieeetr}  
\bibliography{references}

\end{document}